\documentclass[british, twoside]{article}

\usepackage[accepted]{aistats2020}
 \usepackage[nottoc]{tocbibind}
 \usepackage{babel}
\usepackage[utf8]{inputenc}
\usepackage[round]{natbib}

\usepackage{hyperref}
\usepackage{graphicx}
\usepackage{mathrsfs}
\usepackage{amsmath}
\usepackage{amssymb}
\usepackage{algorithm}
\usepackage{algpseudocode}
\usepackage{mathtools}
\usepackage{booktabs}       
\usepackage{subfig}
\usepackage{bm}

\DeclareMathOperator*{\argmax}{argmax}
\DeclareMathOperator{\expect}{\mathbb{E}}
\DeclareMathOperator*{\KL}{KL}
\DeclareMathOperator*{\ent}{\mathcal{H}}
\DeclareMathOperator{\ELBO}{\mathcal{L}}
\DeclareMathOperator*{\N}{\mathcal{N}}

\begin{document}

\twocolumn[

\aistatstitle{Regularising Deep Networks using Deep Generative Models}

\aistatsauthor{  Matthew Willetts \And Alexander Camuto \And  Stephen Roberts \And  Chris Holmes}
\aistatsaddress{ University of Oxford \\ Alan Turing Institute \And University of Oxford \\ Alan Turing Institute \And University of Oxford \\ Alan Turing Institute\And University of Oxford \\ Alan Turing Institute}
]

\begin{abstract}
    We develop a new method for regularising neural networks. We learn a probability distribution over the activations of all layers of the model and then insert imputed values into the network during training. We obtain a posterior for an arbitrary subset of activations conditioned on the remainder. This is a generalisation of data augmentation to the hidden layers of a network, and a form of data-aware dropout. We demonstrate that our training method leads to higher test accuracy and lower test-set cross-entropy for neural networks trained on CIFAR-10 and SVHN compared to standard regularisation baselines: our approach leads to networks with better calibrated uncertainty over the class posteriors all the while delivering greater test-set accuracy. 
\end{abstract}

\section{Introduction}
Methods such a dropout \citep{Srivastava2014}, batch norm \citep{Ioffe2015}, $L_2$ regularisation, data augmentation \citep{deeplearningbook, Wang2017} and ensembling \citep{Lakshminarayanan2017} have been shown to improve generalisation and robustness of deep discriminative models.

We show that by learning a density estimator over the activations of a network during training, and inserting draws from that density estimator into the discriminative model during training, that we get discriminative models with better test set accuracy and better calibration; outperforming all the methods listed above on standard datasets.

Our approach be interpreted as a generalisation of data augmentation to the hidden layers of a network, or from an alternative viewpoint as a from of dropout where we impute activations rather than setting them to $0$.

We specify this density over activations by developing on the ideas of a recent model, VAEAC \citep{Vetrov2019}, a deep generative model (DGM), that enabled the computation of the conditional distribution for arbitrary subsets of pixels of an input image conditioned on the remainder.

After having been trained with imputed values for activations, at test time the discriminative model can either be run as a simple feed-forward model, or, following MC Dropout \citep{Gal2016} we can sample from the model for activations to obtain an estimate of the classifiers uncertainty.

A statistical metric for the quality of the uncertainty of a model is its calibration \citep{Dawid1982}: is a model as likely to be correct in a particular prediction as it is confident in that prediction?
A well-calibrated predictive distribution is key for robust decision making, including in the case of asymmetric losses.
Our proposed form of regularisation leads to better model calibration than standard baselines in pure feed-forward operation and increased test set accuracy.

The key contributions of this paper are:
\begin{itemize}
    \item The introduction of \textit{Pilot} - a model that simultaneously trains a discriminative model and a deep generative model over the former's activations.
    \item Showing that, when applied to both multi-layer perceptron and convolutional neural networks, \textit{Pilot} results in increased accuracy when classifying SVHN and CIFAR-10, beating our baselines.
    \item Demonstrating that the discriminative models trained using samples from our generative model are better calibrated than various baselines.
    \item Finally, showing that when using samples from the model over activations to give model uncertainty, our method outperforms MC-Dropout.
\end{itemize}

\section{Related Work}

Bayesian neural networks (BNNs) \citep{MacKay1992, Neal1995}, where a prior is placed over the parameters of the model and the training data is used to evaluate the posterior over those parameters, give many benefits.
Among them is uncertainty over predictions.
However, as exact inference is not commonly computationally feasible for BNNs, various methods of approximation have been proposed.
These include variational inference \citep{Graves2011, Blundell2015}, expectation propagation \citep{Lobato2015}, and MCMC methods \citep{Welling2011}.
Our approach is analogous to a BNN where we concern ourselves with modelling the discriminative model's activations, not its weights and biases.

Generally, \cite{Poole2014} observed that adding noise, drawn from fixed distributions, to the hidden layers of deep neural networks leads to improved performance.
Our model also has ties to meta-learning, particularly \textit{Hallucination} \citep{Hariharan, Wang2018}, which is an approach to data augmentation in the final hidden layer of a neural network model.
One generates synthetic activations with novel combinations of high-level aspects of the data to represent new or rare classes of data.

Markov chain methods have been developed for data imputation, where a series of draws converges to the underlying data distribution \citep{Bordes2017,Sohl2105,Barber2012}.
\cite{Nazabal2018} extends variational autoencoders (VAEs) to impute missing data, including for discrete, count and categorical data.
There has been interest in using generative adversarial networks \citep{Goodfellow2014} to provide data augmentation \citep{Kiyoiti2019, Antoniou2018, Bowles2018, Fridadar2018} and imputation \citep{Yeh2017, Yoon2018}.

Re-calibrating the probabilities of a discriminative model can enable the construction of a well-calibrated model.
Platt scaling  \citep{Platt1999} and binning methods \citep{Zadrozny2001, Zadrozny2002} are well studied \citep{Niculescu2005}.
Temperature scaling \citep{Jaynes1957} has been shown to produce well-calibrated DNNs \citep{Guo_calib_2017}.

Ensembling of models also leads to better uncertainty in discriminative models \citep{Lakshminarayanan2017,Dietterich2000, Minka2000}.
Dropout \citep{Srivastava2014} can be interpreted as a form of model ensembling.
If dropout is used at test time, as in Monte Carlo (MC) Dropout \citep{Gal2016}, we are in effect sampling sets of models: sub-networks from a larger neural network.
The samples obtained from the predictive distribution provide an estimate of model uncertainty.

Tuning the dropout rate to give good calibration is challenging, and grid search is expensive, which motivates Concrete dropout \citep{Gal_conc_2017} where gradient descent is used to find an optimal value.
The method is strongest for reinforcement learning models, showing lesser performance gains for classification.

\cite{Kingma2015b} show that training a neural network with Gaussian dropout \citep{Wang2013} to maximise a variational lower bound enables the learning of an optimal dropout rate.
However, when the dropout rate is tuned in such a fashion, it is harder to interpret the resulting model as an ensemble \citep{Lakshminarayanan2017}.




\section{Background}

\subsection{Review of VAEAC: VAE with Arbitrary Conditioning}

Briefly we will overview the recent model VAE with Arbitrary Conditioning (VAEAC) \citep{Vetrov2019} - a generalisation of a Conditional VAE \citep{Sohl2105} - as it forms the basis for our approach.
The problem attacked in \citep{Vetrov2019} is dealing with missing data in images via imputation.
They amortise over different arbitrary subsets of pixels, such that training and running the model is relatively cheap.

In \citet{Vetrov2019} there are images $x$ and a binary mask $b$ of unobserved features.
That is, the unobserved data is $x_b$ and the observed data is $x_{1-b}$.
The aim is to build a model, with parameters $\theta$, to impute the value of $x_b$ conditioned on $x_{1-b}$ that closely approximates the true distribution: $p_\theta(x_b | x_{1-b}, b) \approx p(x_b | x_{1-b}, b)$.

Given a dataset $x \sim D$ and a mask prior $p(b)$ we aim to maximise the log likelihood for this problem wrt $\theta$:
\begin{equation}
    \theta^* = \argmax_\theta \expect_{x\sim D} \expect_{b\sim p(b)} \log p_\theta(x_b | x_{1-b}, b)
\end{equation}
Introducing a continuous latent variable $z$ gives us the VAEAC generative model:
\begin{equation}
    p_\theta(x_b | x_{1-b}, b) = \int \mathrm{d}z \, p_\theta(x_b | z, x_{1-b}, b) p_\theta(z|x_{1-b}, b)
\end{equation}
Where $p_\theta(z|x_{1-b}, b) = \N(z|\mu_\theta(x_{1-b},b),\Sigma_\theta(x_{1-b},b))$, and $p_\theta(x_b | z, x_{1-b}, b)$ is an appropriate distribution for the data $x$.
The parameters of both are parameterised by neural networks.

By introducing a variational posterior $q_\phi(z|x, b) = \N(z|\mu_\phi(x,b),\Sigma_\phi(x,b))$ and obtain the VAEAC evidence lower bound (ELBO) for a single data point and a given mask:
\begin{align}
    \ELBO^{\mathrm{VAEAC}}(x, b; \theta, \phi) =& \expect_{z\sim q}\log  p_\theta(x_b | z, x_{1-b}, b) \\
    &- \KL(q_\phi(z|x, b) || p_\theta(z|x_{1-b}, b)) \nonumber
\end{align}

Note that the variational posterior $q$ is conditioned on all $x$, so that in training under this objective we must have access to complete data $x$.
When training the model they transfer information from $q_\phi(z|x, b)$ which has access to all $x$ to the $p_\theta(z|x_{1-b}, b)$ that does not, by penalising the $\KL$ divergence between them.
At test time, when applying this model to real incomplete data, they sample from the generative model to infill missing pixels.


\subsection{Training Classifiers with Data Augmentation and Dropout}

We wish to train a discriminative model $p_\Psi(y|x)$, where $y$ is a the output variable,  $x$ an input (image), and $\Psi$ are the parameters of the network.
We focus here on classification tasks, where in training we aim to minimise the cross-entropy loss of the network, or equivalently maximise the log likelihood of the true label under the model, $\ELBO(D; \Psi)$ wrt $\Psi$ for our training data $(x^*,y^*) \sim D$:
\begin{align}
    \ELBO(D;\Psi) & = \expect_{(x,y) \sim D} \log p_\Psi(y|x) \\
    \Psi^* &= \argmax_\Psi \ELBO(D; \Psi)
\end{align}

Commonly one might a train the model with methods like dropout or data augmentation to regularise the network.
Here we consider data augmentation as a probabilistic procedure, and write out (MC) dropout in similar notation.
Then we describe our approach to learning a density estimator over activations of a DNN, which we then use as a generalisation of both data augmentation and dropout in training regularised deep nets.

\subsubsection{Data Augmentation}

If we have a discriminative classifier $p_\Psi(y|x)$, we could train it on augmented data $\tilde{x}$.
If the procedure for generating the augmentation is stochastic we could represent it as $p_\theta(\tilde{x}|x)$.
This could correspond, say, to performing transformations (like rotating or mirroring) on some proportion $\theta$ of each batch during training.
Thus we can write the joint distribution for the classifier and the `augmenter' $p_\theta(\tilde{x}|x)$, conditioned on $x$, as:
\begin{equation}
    p_{\Psi, \theta}(y, \tilde{x} | x) = p_\Psi(y|\tilde{x})p_\theta(\tilde{x}|x) \label{eq:data_aug_prob}
\end{equation}
And so marginalising out the augmenter:
\begin{equation}
    p_{\Psi, \theta}(y|x) = \expect_{\tilde{x} \sim p_\theta(\tilde{x}|x)} p_\Psi(y|\tilde{x}) \label{eq:data_aug_marginal}
\end{equation}

\subsubsection{Dropout}

Taking a probabilistic perspective to $\Psi$, the weights and biases of the network, we would write the same classifier as $p(y|x, \Psi)$.
A manipulation of the weights by a stochastic method, such as dropout, can be written as $p_\theta(\tilde{\Psi}|\Psi)$.
For dropout, $\theta$ would be the dropout rate.
So the equivalent to Eq (\ref{eq:data_aug_prob}) is:
\begin{equation}
    p_\theta(y, \tilde{\Psi}|x, \Psi) = p(y|x, \tilde{\Psi}) p_\theta(\tilde{\Psi}|\Psi) \label{eq:dropout_prob}
\end{equation}
And to Eq (\ref{eq:data_aug_marginal}):
\begin{equation}
    p_{\Psi,\theta}(y|x) = \expect_{\tilde{\Psi} \sim p_\theta(\tilde{\Psi}|\Psi)} p(y|x, \tilde{\Psi}) \label{eq:dropout_marginal}
\end{equation}

\subsubsection{Loss functions}
For both data augmentation and dropout, the model is still trained on the expected log likelihood of the output variable, but is now being fed samples from the data augmentation pipeline or dropout mask.
We can obtain this loss by applying Jensen's inequality to the logarithms of Eqs (\ref{eq:data_aug_marginal}, \ref{eq:dropout_marginal}) and taking expectations over $D$:
\begin{align}
    \ELBO^{\mathrm{aug}}(D;\Psi,\theta) =& \expect_{x,y\sim D}\expect_{\tilde{x} \sim p_\theta(\tilde{x}|x)} \log p_\Psi(y|\tilde{x})\\
    \ELBO^{\mathrm{drop}}(D;\Psi,\theta) =& \expect_{x,y\sim D}\expect_{\tilde{\Psi} \sim p_\theta(\tilde{\Psi}|\Psi)} \log p(y|x, \tilde{\Psi})
\end{align}
Training is then done by maximising $\ELBO^{\mathrm{aug}}$ or $\ELBO^{\mathrm{drop}}$ wrt $\Psi$.
In principle $\theta$ may contain other parameters of the data augmentation or dropout procedure, which could be learnt jointly, but are commonly fixed.

\subsection{Stochastic manipulation of activations}

We wish to obtain a conditional distribution for any subset of activations, conditioned on the remaining activations.
Consider our discriminative model as being composed of $L$ layers.
We view our input data and the activations of the network on an equal footing; the output of one layer acts as input to the next layer, and can be viewed as equivalent to data for that later layer. 
In this view the input data to the discriminative model, $x$, is simply the $0^{\mathrm{th}}$ layer of activations.
We record the read out of the activations of every unit of every layer in the model $p_\Psi(y|x)$ for a given data-point, namely: 
\begin{equation}
    a = f_\Psi(x)
\end{equation}
where $a^0 = x$ and $\mathrm{softmax}(a^L)=p_\Psi(y|x)$.

In analogy with the section above we denote a stochastic procedure for generating different realisations of the activations $\tilde{a}$ given the recorded activations $a$ as $p_\theta(\tilde{a}|a)$.
The joint is then:
\begin{equation}
    p_{\Psi,\theta}(y, \tilde{a}|a) = p_\Psi(y|\tilde{a}) p_\theta(\tilde{a}|a) \label{eq:pilot_prob}
\end{equation}
Marginalising out $\tilde{a}$ and taking the logarithm we obtain:
\begin{equation}
    \log p_{\Psi, \theta}(y|a) = \log \expect_{\tilde{a} \sim p_\theta(\tilde{a}|a)}[p_\Psi(y|\tilde{a})] \label{eq:pilot_marginal_one_datapoint}
\end{equation}
Applying Jensen's Inequality:
\begin{equation}
     \log p_{\Psi, \theta}(y|a) \geq \expect_{\tilde{a} \sim p_\theta(\tilde{a}|a)}[\log p_\Psi(y|\tilde{a})]
\end{equation}
And taking an expectation over the dataset $D$:
\begin{equation}
    \ELBO^\mathrm{act}(D;\Psi,\theta) = \expect_{\substack{(x,y) \sim D \\ a=f_\Psi(x)}} \expect_{\tilde{a} \sim p_\theta(\tilde{a}|a)}[\log p_\Psi(y|\tilde{a})] \label{eq:pilot_marginal}
\end{equation}
$\ELBO^\mathrm{act}$ is our classification objective.
We will optimise it wrt $\Psi$, not taking its gradient wrt $\theta$.
In the next section we introduce a particular form for $p_\theta(\tilde{a}|a)$, which we will then learn simultaneously.

\section{Pilot: DGM over Activations}

As well as training the model parameters $\Psi$ when our model is being run with imputed activations $\tilde{a}$, we also wish for our model to generate realistic activations $\tilde{a}$.

To learn a density estimator over activations, we define a parametric generative model for $p_\theta(\tilde{a}|a)$ that we will then train using amortised stochastic variational inference \citep{Kingma2013, Rezende2014}.

In analogy to the image in-painting of \cite{Vetrov2019}, we impute a subset of a network's activations given the values of the remainder.
Likewise, we introduce a mask $b$ with prior $p(b)$ so $\tilde{a}_b$ are the values we will impute given the unmasked variables $a_{1-b}$.

That means we choose:
\begin{equation}
     p_\theta(\tilde{a}|a) = \expect_{b\sim p(b)} p_\theta(\tilde{a}_b|a_{1-b},b)
     \label{eq:a_form}
\end{equation}
where $\tilde{a}$ is constructed deterministically by taking the values $\tilde{a}_b$ in the masked positions with the recorded values $a_{1-b}$. We denote this as a form of masked element-wise addition:
\begin{equation}
    \tilde{a} = \tilde{a}_b \oplus a_{1-b}
\end{equation}
We wish to train the model $p_\theta(\tilde{a}_b|a_{1-b},b)$ so that $\tilde{a}_b$ is close to the real activations $a_b$.
As in \cite{Vetrov2019} we introduces a latent variable $z$, defining the generative model as:
\begin{align}
    p_\theta(\tilde{a}_b|a_{1-b},b) &= \int \mathrm{d}z \, p_\theta(\tilde{a}_b|a_{1-b},b,z)p_\theta(z|a_{1-b},b) \label{eq:log_expect1} \\
     &= \expect_{z \sim p_\theta(z|a_{1-b},b)} p_\theta(\tilde{a}_b|a_{1-b},b,z)
\end{align}

Where $p_\theta(z|a_{1-b},b) = \N(z|\mu_\theta(a_{1-b},b),\Sigma_\theta(a_{1-b},b))$ in analogy with VAEAC.
Our aim is to maximise the log likelihood:
\begin{equation}
\expect_{\substack{(x,y) \sim D \\ a=f_\Psi(x)}}\expect_{b\sim p(b)}\log p_\theta(a_b|a_{1-b},b)
\label{eq:loglike}
\end{equation}

To train this model, we introduce a variational posterior for $z$, which is conditioned on all $a$:
$q_\phi(z|a, b) = \N(z|\mu_\phi(a,b),\Sigma_\phi(a,b))$.
This is unlike its generative counterpart $p_\theta(z|a_{1-b},b)$ which only receives the masked activity information $a_{1-b}$.
This gives us an ELBO for $\log p_\theta(\tilde{a}_b|a_{1-b},b)$ which we denote $\Lambda$:
\begin{align}
    \Lambda(a, b;\theta,\phi) =& \expect_{z\sim q}[\log p_\theta(a_b|a_{1-b},b,z)] \notag \\
    &- \KL(q_\phi(z|a,b)||p_\theta(z|a_{1-b},b)) \label{eq:lambda} \\
    \ELBO^{\mathrm{DGM}}(D;\theta,\phi) =& \expect_{\substack{(x,y)\sim D \\ b \sim p(b)}}[\Lambda(a=f_\Psi(x),b;\theta,\phi)] \label{eq:pilot_elbo}
\end{align}
As stated previously, this objective leads to information being passed from $q_\phi(z|a,b)$, the part of the model that can access all $a$, to $p_\theta(z|a_{1-b},b))$ the part of the model that only sees the masked activation values $a_{1-b}$.

We choose to model the raw activations of the model, before applying an activation function.
This sidesteps the potential difficulties if we modelled them after the application of an activation function - for instance applying ReLUs' gives us values that are $>0$.
We model our raw activations with Gaussian likelihood $\log p_\theta(a_b|a_{1-b},b,z)$ with fixed diagonal covariance.
We place a Normal-Gamma hyperprior on $p_\theta(z|a_{1-b},b)$ to prevent large means and variances. See Appendix \ref{app:hyperprior} for a full definition.

\subsection{Overall Objective}


To train both the DGM (that produces samples for $a_b$) and classifier we maximise their objectives simultaneously:
\begin{align}
    \ELBO^\mathrm{pilot}(D;\Psi,\theta, \phi) =&  \ELBO^\mathrm{act}(D;\Psi) + \ELBO^{\mathrm{DGM}}(D;\theta,\phi) \label{eq:overall}
\end{align}
We train the model by optimising $\ELBO^\mathrm{act}$ wrt $\Psi$ while simultaneously optimising $\ELBO^{\mathrm{DGM}}$ wrt $\theta, \phi$, both by stochastic gradient descent using Adam \citep{Kingma2015} over $D$.
The objective $\ELBO^{\mathrm{DGM}}$ could be written as a function of $\Psi$ as well, as $a=f_\Psi(x)$, but we choose not to take gradients wrt $\Psi$ through $\ELBO$, 
similarly for $\ELBO^\mathrm{act}$ and $\theta,\phi$.

This separation is key to the proper functioning of our model.
If we optimised $\ELBO^\mathrm{act}$ wrt $\theta,\phi$ then the larger, more powerful DGM would perform the task of the classifier - the DGM could learn to simply insert activations that gave a maximally clear signal that a simplistic classifier could then use. The classifier could then fail to operate in the absence of samples $\tilde{a}$, and the DGM would be the real classifier.
If we optimised $\ELBO^\mathrm{DGM}$ wrt $\Psi$, we would in effect be training the discriminator to be more amenable to being modelled by the DGM.
An interesting idea perhaps, but a different kind of regularisation to that which we wish to study.

We take MC samples to approximate the integrals in $\ELBO^{\mathrm{DGM}}$, employing the reparameterisation trick to take differentiable samples from our distributions \citep{Kingma2013, Rezende2014}.

\section{Calibration Metrics}

A neural network classifier gives a prediction $\hat{y}(x)$ with confidence $\hat{p}(x)$ (the probability attributed to that prediction) for a datapoint $x$.
Perfect calibration consists of being as likely to be correct as you are confident:
\begin{equation}
    p(\hat{y}=y|\hat{p}=r)=r, \quad \forall r\in[0,1]
\end{equation}
To see how closely a model approaches perfect calibration, we plot reliability diagrams \citep{degroot1983, Niculescu2005}, which show the accuracy of a model as a function of its confidence over $M$ bins $B_m$.
\begin{align}
    \mathrm{acc}(B_m) &= \frac{1}{|B_m|}\sum_{i\in B_m} \mathcal{1}(\hat{y}_i = y_i)\\
    \mathrm{conf}(B_m) &= \frac{1}{|B_m|}\sum_{i\in B_m} \hat{p}_i
\end{align}
We also calculate the Expected Calibration Error (ECE) \cite{Naeini2015}, the mean difference between the confidence and accuracy over bins:
\begin{equation}
    \mathrm{ECE} = \sum_{m=1}^M \frac{|B_m|}{N}|\mathrm{acc}(B_m) - \mathrm{conf}(B_m)|
    \label{eq:ece}
\end{equation}
However, ECE is not a perfect metric.
With a balanced test set one can trivially obtain ECE $\approx 0$ by sampling predictions from a uniform distribution over classes. Nevertheless, ECE is a valuable metric in conjunction with a model's reliability diagram.

\section{Experiments}\label{sec:experiments}

\begin{table*}
\vspace{1em}
  \caption{Test set accuracy, mean per-datapoint negative log-likelihood ($\mathrm{NLL}= -\ELBO_{\mathrm{xent}}$) and ECE [see Eq (\ref{eq:ece})] for convolutional neural networks and 2-hidden-layer MLPs (with 1024 hidden units) trained on CIFAR-10 and SVHN with different regularisation methods}
  \label{table:results}
  \centering
    \setlength\tabcolsep{1.9pt} 
  \begin{tabular}{rcccccc}
    \toprule
    \multicolumn{7}{c}{\textbf{CNN models}}\\
    \toprule    Model & Acc$(D^\mathrm{CIFAR10}_\mathrm{test})$ & Acc$(D^\mathrm{SVHN}_\mathrm{test})$& $\mathrm{NLL}(D^\mathrm{CIFAR10}_\mathrm{test})$ &  $\mathrm{NLL}(D^\mathrm{SVHN}_\mathrm{test})$ & ECE$(D^\mathrm{CIFAR10}_\mathrm{test})$ & ECE$(D^\mathrm{SVHN}_\mathrm{test})$ \\
    \midrule
    Vanilla                 & $0.630 \pm 0.003$ & $ 0.846 \pm 0.001$  & $3.54 \pm 0.05$ & $ 1.69 \pm 0.00$ & $0.308 \pm 0.003$ & $0.122 \pm 0.002$\\
    \midrule
    Pilot \textit{a-aug}    & $\bm{0.701 \pm 0.005}$ & $\bm{0.881 \pm 0.005}$ & $\bm{0.87 \pm 0.02}$ & $\bm{0.44 \pm 0.00}$ & $\bm{0.012 \pm  0.000}$ & $0.033 \pm  0.002$\\
    Pilot \textit{a-drop}   & $0.454 \pm 0.035$ & $0.200 \pm 0.001$  & $ 1.59 \pm 0.01$ & $2.29 \pm 0.01$ & $0.096 \pm  0.010$ & $0.092 \pm 0.002$  \\
    Pilot \textit{x-aug}    & $0.648 \pm 0.001$ & $0.861 \pm 0.001$ & $1.49 \pm 0.01$ & $ 0.64 \pm 0.002$ & $0.210 \pm  0.005$ & $0.066 \pm  0.001$\\
    Pilot \textit{x-drop}   & $0.625 \pm 0.04$ & $0.844 \pm 0.002$ & $1.15 \pm 0.01$ & $0.55 \pm 0.01$ & $0.116 \pm 0.002$ & $0.019 \pm  0.001$\\

    \midrule
    Add  \textit{a-aug}     & $0.641 \pm 0.001$ & $0.858 \pm 0.012$  & $4.80 \pm 0.02$ & $ 1.05 \pm 0.06$ & $0.199 \pm 0.000$ & $0.065 \pm 0.012$ \\
    Add  \textit{a-drop}    & $0.630 \pm 0.002$ & $0.850 \pm 0.011$  & $2.05 \pm 0.02$ & $ 0.89 \pm 0.12$ & $0.249 \pm 0.000$ & $0.081 \pm 0.001$ \\
    Add  \textit{x-drop}    & $0.609 \pm 0.011$ & $0.844 \pm 0.000$  & $1.44 \pm 0.00$ & $ 0.56 \pm 0.01$ & $0.184 \pm 0.002$ &  $0.033 \pm 0.001$ \\
     Sub  \textit{a-drop}   & $0.403 \pm 0.001$ & $0.748 \pm 0.070$ & $1.43 \pm 0.00$ & $ 1.52 \pm 0.02$ & $0.490 \pm 0.001 $ & $0.155 \pm 0.001$ \\
    Sub  \textit{x-drop}    & $0.521 \pm 0.086$ & $ 0.742 \pm 0.007$  & $1.97\pm 0.05$ & $ 0.88 \pm 0.04$ & $0.199 \pm 0.001$ & $0.032 \pm 0.001$\\

    \midrule
    Dropout                 & $0.629 \pm 0.002$ & $ 0.850 \pm 0.001$  & $3.57  \pm 0.01$&$ 1.68 \pm 0.01$ & $0.308 \pm 0.001$ & $0.121 \pm 0.001$\\
    $L_2$, $\lambda=0.1$    & $0.629 \pm 0.002$ & $ 0.847 \pm 0.000$  & $3.59 \pm 0.05$ & $ 1.69 \pm 0.00$ & $0.308 \pm 0.004$ & $0.123 \pm 0.001$\\
    Batch norm              & $0.631 \pm 0.001$ & $ 0.846 \pm 0.001$  & $4.60 \pm 0.02$ & $ 2.12 \pm 0.02$ & $0.230 \pm 0.010 $ & $0.054 \pm 0.000$\\
    Data Aug                & $0.646 \pm 0.001$ & $ 0.750 \pm 0.002$  & $1.027 \pm 0.00$ & $ 0.77 \pm 0.01$ & $0.016 \pm 0.002$ & $\bm{0.009 \pm 0.001}$\\
    \midrule
    \midrule
    Pilot$_{\mathrm{MC}}$ \textit{a-aug} & $0.700 \pm 0.002$ & $0.877 \pm 0.002$ & $0.94 \pm 0.01$ & $0.53 \pm 0.00$ & $0.089 \pm 0.001$ & $0.120 \pm 0.001$  \\
    Pilot$_{\mathrm{MC}}$ \textit{a-drop} & $0.453 \pm 0.002$ & $0.196 \pm 0.000$ & $1.57 \pm 0.02$ & $2.25 \pm 0.00$ & $0.065 \pm 0.003$ & $0.087 \pm 0.001$ \\
    \midrule
    Add$_{\mathrm{MC}}$ \textit{a-aug}     & $0.576 \pm 0.035$ & $0.860 \pm 0.001$ & $1.67 \pm 0.01$ & $0.56 \pm 0.00$ & $0.063 \pm 0.001$ & $0.017 \pm 0.002$ \\
    Add$_{\mathrm{MC}}$ \textit{a-drop} & $0.636 \pm 0.020$ & $0.854 \pm 0.001$ & $1.73 \pm 0.01$ & $0.73 \pm 0.00$ & $0.199 \pm 0.001$ & $0.0528 \pm 0.001$ \\
    \midrule
    MC Dropout              & $0.579 \pm 0.001$ & $ 0.795 \pm 0.002$  & $1.69  \pm 0.01$&$ 0.93 \pm 0.00$ & $0.065 \pm 0.000$ & $0.067 \pm 0.005$ \\
    Ensemble                & $0.683 \pm 0.001$ & $ 0.870 \pm 0.001$  & $0.96 \pm 0.01$ & $0.51 \pm 0.01$ & $0.025 \pm 0.003$ & $0.060 \pm 0.002$ \\
    \bottomrule
    \toprule
    \multicolumn{7}{c}{\textbf{MLP models}}\\
    \toprule
    Model & Acc$(D^\mathrm{CIFAR10}_\mathrm{test})$ & Acc$(D^\mathrm{SVHN}_\mathrm{test})$& $\mathrm{NLL}(D^\mathrm{CIFAR10}_\mathrm{test})$ &  $\mathrm{NLL}(D^\mathrm{SVHN}_\mathrm{test})$ & ECE$(D^\mathrm{CIFAR10}_\mathrm{test})$ & ECE$(D^\mathrm{SVHN}_\mathrm{test})$ \\
    \midrule
     Vanilla                     & $0.581 \pm 0.003$ & $ 0.848 \pm 0.001$  & $4.78 \pm 0.03$ & $ 2.17 \pm 0.06$
     &$0.470 \pm 0.004$&$0.127 \pm 0.003$ \\
    \midrule
    Pilot \textit{a-aug}         & $\bm{0.601 \pm 0.001}$ & $\bm{0.858 \pm 0.002}$ & $\bm{1.22 \pm 0.01}$ & $\bm{0.53 \pm 0.02}$ 
    & $0.056 \pm 0.004$ &$\bm{0.014} \pm 0.001$\\
    Pilot \textit{a-drop}        & $0.517 \pm 0.001$ & $0.794 \pm 0.002$ & $1.36 \pm 0.01$ & $ 0.79 \pm 0.02$ 
    & $0.110 \pm 0.003$ &$0.029 \pm  0.002$\\
    Pilot \textit{x-aug}         & $0.565 \pm 0.002$ & $0.851 \pm 0.001$ & $2.42 \pm 0.01$ & $ 1.16 \pm 0.00$ 
    & $0.288 \pm 0.002$ &$0.057 \pm  0.002$\\
    Pilot \textit{x-drop}        & $0.570 \pm 0.002$ & $0.837 \pm 0.003$ & $2.14 \pm 0.07$ & $0.72 \pm 0.001$ 
    & $0.284 \pm 0.017$ &$0.057 \pm 0.001$\\

    \midrule
    Add  \textit{a-aug}         & $0.578 \pm 0.001$ & $0.843 \pm 0.001$  & $2.76 \pm 0.01$ & $ 0.78 \pm 0.06$ 
    & $0.301 \pm 0.000$ &$0.077 \pm 0.001$ \\
    Add  \textit{a-drop}     & $0.578 \pm 0.004$ & $0.849 \pm 0.031$  & $4.26 \pm 0.02$ & $ 1.48 \pm 0.12$
    & $0.345 \pm 0.002$&$0.114 \pm  0.004$\\
    Add  \textit{x-drop}     & $0.547 \pm 0.043$ & $0.841 \pm 0.000$  & $2.99 \pm 0.01$ & $ 0.75 \pm 0.01$
    & $0.307 \pm 0.001$ &$0.067 \pm 0.002$\\
     Sub  \textit{a-drop}    & $0.462 \pm 0.041$ & $0.737 \pm 0.079$ & $4.23 \pm 1.23$ & $ 1.92 \pm 0.02$
     & $0.403 \pm 0.131$ &$0.143 \pm  0.001$\\
    Sub  \textit{x-drop}     & $0.499 \pm 0.002$ & $ 0.765 \pm 0.001$  & $2.22\pm 0.02$ & $ 0.80 \pm 0.03$
    &$0.279 \pm 0.001$&$0.029 \pm  0.003$\\

    \midrule
    Dropout           & $0.570 \pm 0.002$ & $ 0.837 \pm 0.049$  & $4.88  \pm 0.01$&$ 1.27 \pm 0.24$
    &$0.480 \pm 0.001$&$0.116 \pm 0.021$\\
    $L_2$, $\lambda=0.1$        & $0.574 \pm 0.002$ & $ 0.847 \pm 0.000$  & $4.74 \pm 0.02$ & $ 2.12 \pm 0.00$
    &$0.479 \pm 0.001$&$0.127 \pm 0.001$\\
    Batch norm                  & $0.579 \pm 0.001$ & $ 0.848 \pm 0.001$  & $4.55 \pm 0.02$ & $ 2.04 \pm 0.02$
    &$0.570 \pm 0.001$&$0.162 \pm 0.002$\\
    Data Aug                    & $0.566 \pm 0.001$ & $ 0.731 \pm 0.001$  & $1.36 \pm 0.01$ & $ 0.91 \pm 0.01$
    & $0.231 \pm 0.001$&$0.055 \pm 0.001$\\
    \midrule
    \midrule
    Pilot$_{\mathrm{MC}}$ \textit{a-aug}     & $0.598 \pm 0.002$ & $0.855 \pm 0.001$ & $1.24 \pm 0.01$ & $0.56 \pm 0.00$ 
    &$0.036 \pm 0.001$&$0.042 \pm 0.002$\\
    Pilot$_{\mathrm{MC}}$ \textit{a-drop}    & $0.519 \pm 0.001$ & $0.761 \pm 0.002$ & $1.37 \pm 0.01$ & $0.84 \pm 0.01$ 
    &$0.066 \pm 0.002$&$0.107 \pm 0.003$\\
    \midrule
    Add  \textit{a-aug}         & $0.576 \pm 0.003$ & $0.839 \pm 0.001$ & $2.65 \pm 0.02$ & $0.73 \pm 0.00$ 
    &$0.296 \pm 0.001$&$0.056 \pm 0.001$\\
    Add  \textit{a-drop}     & $0.583 \pm 0.023$ & $0.847 \pm 0.001$ & $3.70 \pm 0.01$ & $1.12 \pm 0.00$ 
    &$0.335 \pm 0.003$&$0.087 \pm 0.000$\\
    \midrule
    MC Dropout         & $0.509 \pm 0.002$ & $ 0.784 \pm 0.001$  & $2.19  \pm 0.01$&$ 1.07 \pm 0.01$
    &$0.085 \pm 0.002$&$0.080 \pm 0.004$\\
    Ensemble                    & $0.518 \pm 0.002$ & $ 0.850 \pm 0.001$  & $1.58  \pm 0.01$&$ 1.68 \pm 0.01$
    &$\bm{0.027} \pm 0.002$&$0.155 \pm 0.001$\\
    \bottomrule
  \end{tabular}
\end{table*}

\begin{figure*}
    \centering
    \subfloat[CIFAR10 - CNN]{{ \includegraphics[width=0.7\textwidth]{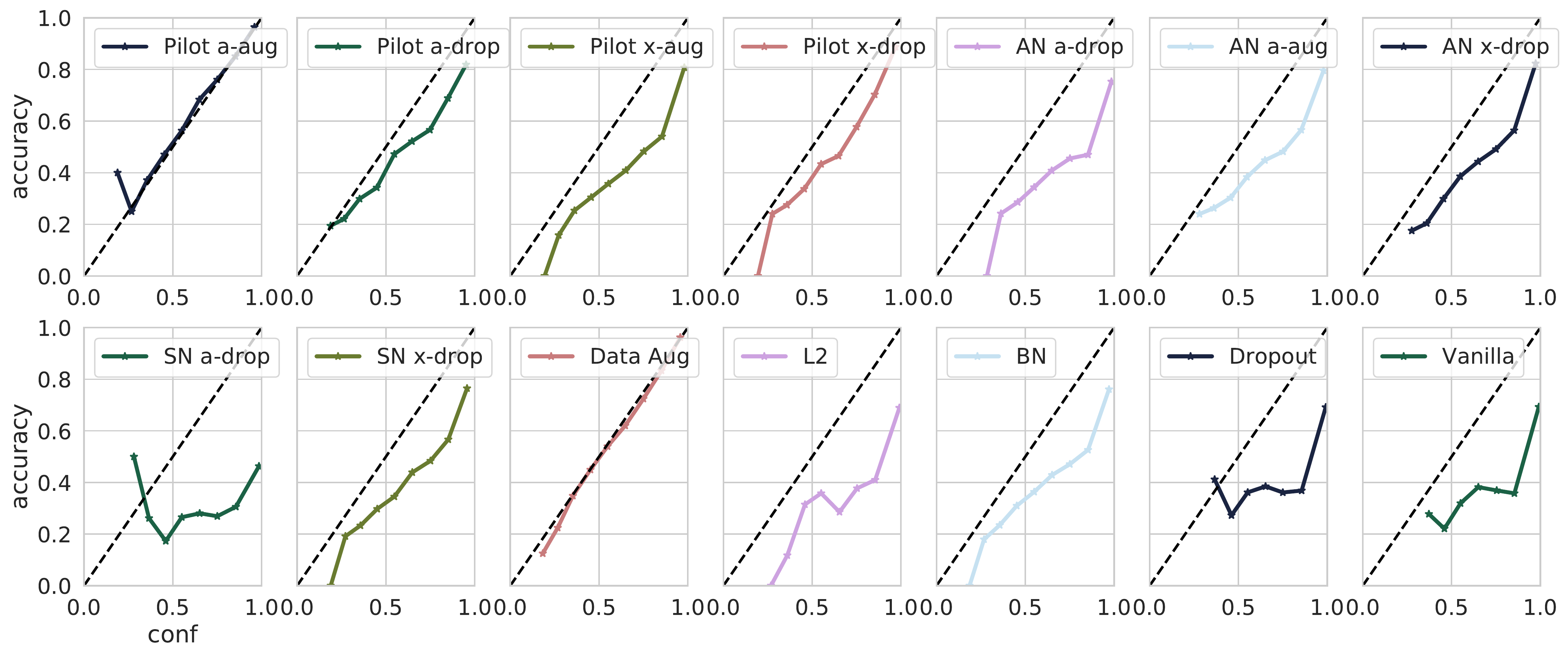} }}%
    \\
    \subfloat[SVHN - CNN]{{
    \includegraphics[width=0.7\textwidth]{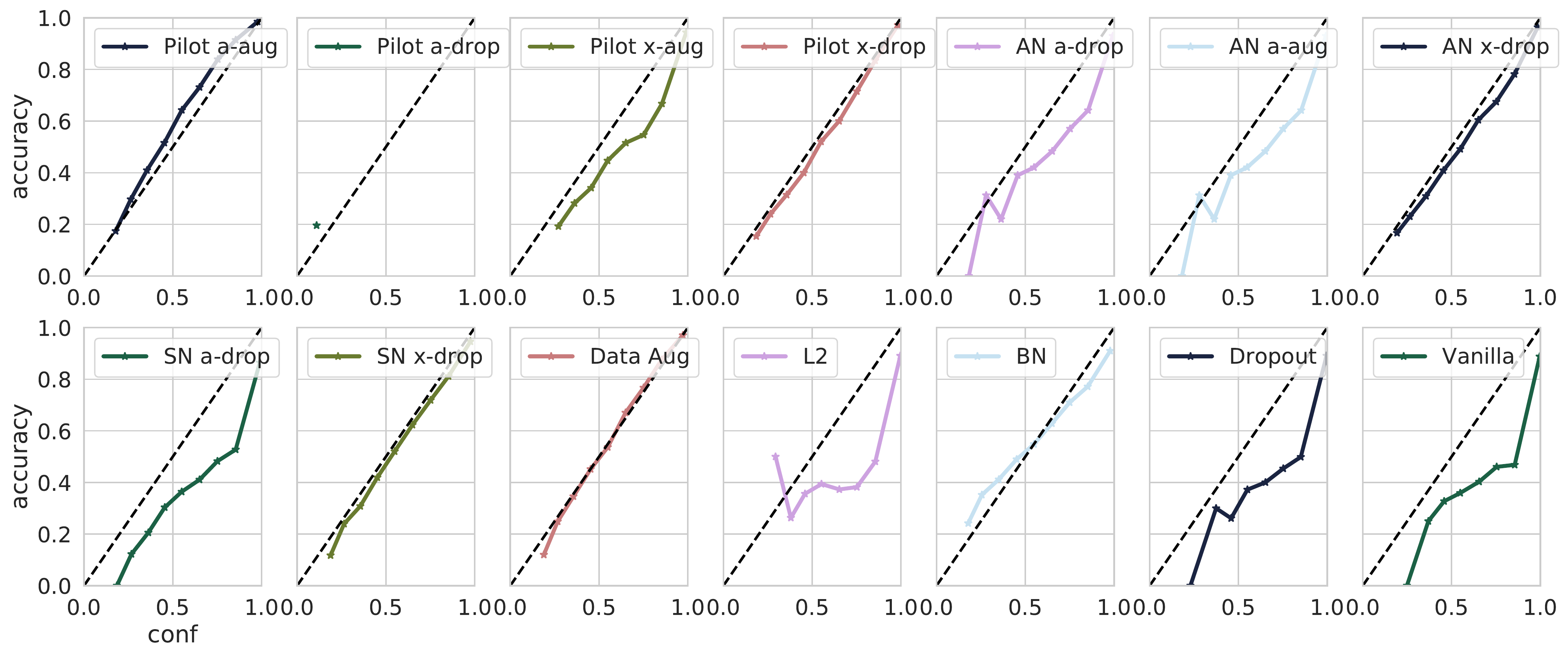} }}%
    \\
    \subfloat[CIFAR10 - MLP]{{ \includegraphics[width=0.7\textwidth]{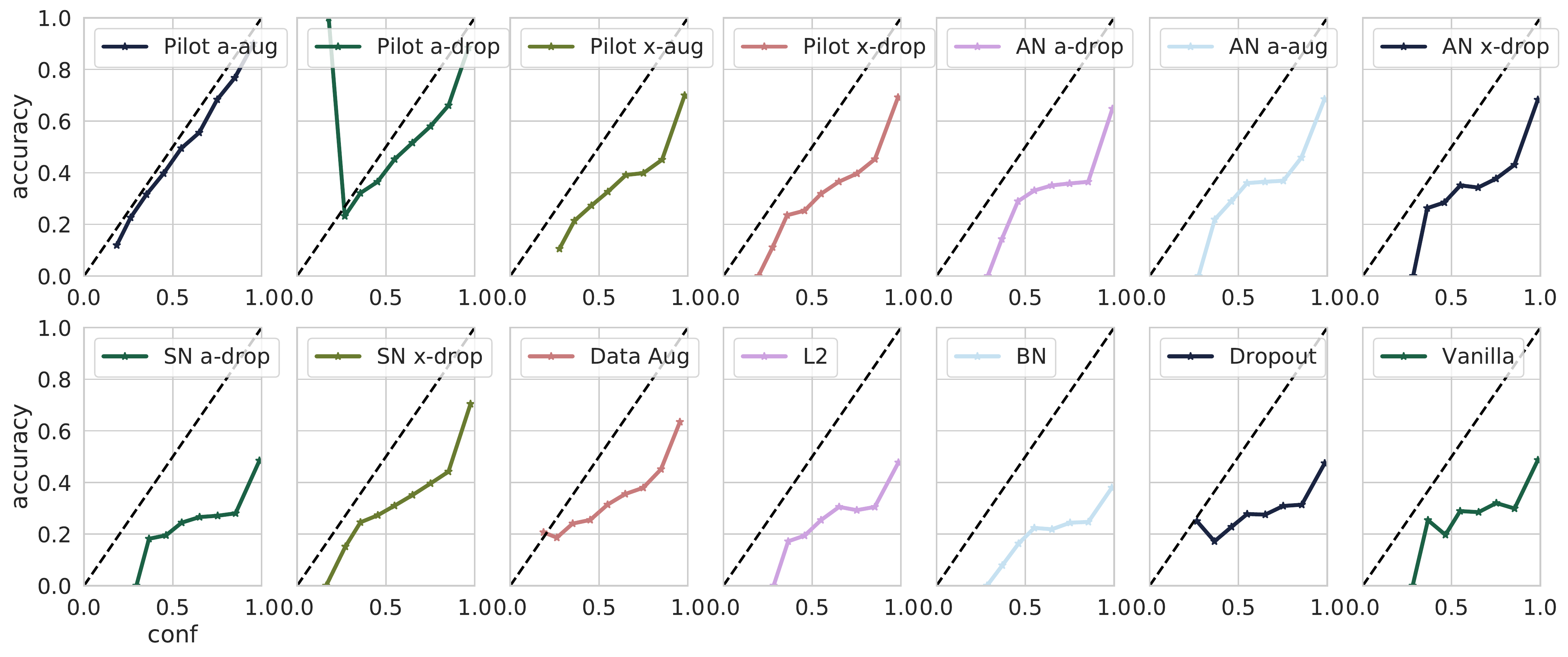}}}%
    \\
    \subfloat[SVHN - MLP]{{ \includegraphics[width=0.7\textwidth]{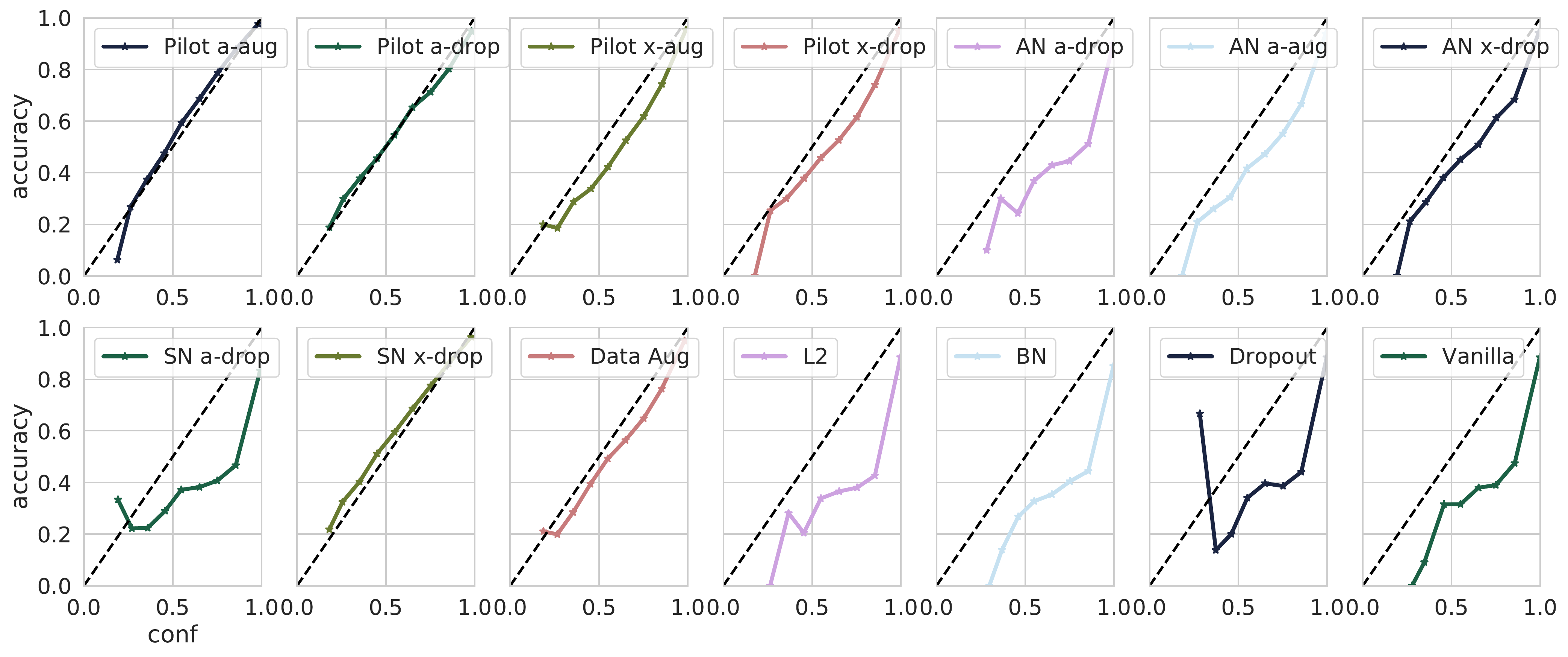} }}
    \caption{Reliability diagrams for SVHN and CIFAR10 test sets for different regularisation methods.}
    \label{fig:calib_plot}
\end{figure*}

We wish to test if our method produces a trained deep net with better calibration, as measured by reliability diagrams, ECE and test-set log likelihood, while maintaining or increasing test-set accuracy.
We benchmark against standard methods to regularise deep nets:
dropout \citep{Srivastava2014} with rate $r=0.5$, batch norm \citep{Ioffe2015} with default parameters, $L_2$ regularisation with weighting $\lambda=0.1$ \citep{deeplearningbook}, and a data augmentation strategy where we introduce colour shifts, rotations and flips to data with probability of $0.1$ for each datapoint.

We propose two modes of operation for \textit{Pilot}. The first, at test-time, simply evaluates $p(y|x)$ without draws from $\tilde{a}$. The second, which we call  Pilot$_{\mathrm{MC}}$, samples numerous realisations of $p(y|x)$ for our model by repeatedly drawing $\tilde{a}$ thus giving uncertainty estimates for predictions.
As such, we also benchmark against methods shown to given model uncertainty estimates for deep nets: ensembles \citep{Lakshminarayanan2017} and MC dropout \citep{Gal2016}. Here we build an ensemble by uniformly weighting the predictions of all benchmarks.
Where uncertainty estimates are obtained by sampling, such as in \textit{Pilot}, MC Dropout, and our noise baselines, we draw 10 samples from the models at test time and average their outputs to generate a prediction.

In addition to standard baselines, we compare our method to a `noisy substitution' (Sub) method where during training, we substitute values of $a$ with draws from $N(0,\sigma^{2})$; and to a `noisy addition' (Add) method whereby we sum noisy draws from $N(0,\sigma^{2})$ with $a$. 
These methods are applied on neurons masked by a mask $b$ as in \textit{Pilot}.
In both cases $\sigma^{2}$ is the same fixed variance as in the DGM's decoder (see below).
Both these methods are linked to \textit{Pilot}, but are also reminiscent of the work in \citep{Poole2014}. Additive noise corresponds to the asymptote of the DGM where it perfectly infers a neuron's activation with a fixed variance $\sigma^{2}$. Substitutive noise corresponds to earlier stages of training where samples are drawn from an uninformative prior. 

We apply our method and benchmarks to 2-hidden-layer multi-layer perceptrons (MLPs) and small convolutional networks, on CIFAR-10 \citep{Krizhevsky2009} and SVHN \citep{SVHN}.
We train the models containing MLPs for 250 epochs and models containing CNNs for 100 epochs. The Appendix includes a subset of the experiments for a smaller MLP. 

We run Pilot in two broad modes: \textit{aug} where we impute a single layer at a time, and; \textit{drop} where, akin to dropout, we randomly sample nodes from across the network.
This leads us to choose four settings for our mask prior $p(b)$, which is applied to Pilot and to the noisy addition and substitution benchmarks.
\begin{itemize}
\itemsep1em 
\item[1)] \textit{$x$ dropout (x-drop)}: $p(b)$ an iid Bernoulli distributions with trial success $r$ over just the input layer $a^0=x$, never masking $a^{\ell>0}$.
\item[2)] \textit{$x$ augment (a-aug)}: we impute all of $a^0=x$ given the other activations, but only for a proportion $r$ during training. 
\item[3)] \textit{activation dropout (a-drop)}: $p(b)$ is a set of iid Bernoulli distributions with trial success $r$ over all units of $p_\Psi(y|x)$.
\item[4)] \textit{activation augment (a-aug)}: we impute all of one layer $a^\ell$ chosen uniformly at random, but only for a proportion $r$ during training.
\end{itemize}


All networks in the DGM parts of our model are MLPs and we fix the variance of our decoder distribution, $ p_\theta(a_b|a_{1-b},b,z)$, a Gaussian with parameterised mean, to 0.1.

\subsection{Results}

From Table \ref{table:results} we can see that Pilot activation augmentation (\textit{a-aug}) leads to better test set accuracy and test set negative log likelihood (NLL) relative to all other models including a vanilla classifier, which has no regularisation applied during training. 

Pilot \textit{a-aug} and Pilot$_{\mathrm{MC}}$ \textit{a-aug} consistently provide low ECE but do not always generate the best calibrated models.
Better calibrated models are generated for SVHN when CNNs are trained with data augmentation (Pilot \textit{a-aug}: ECE=$3.3\%$ vs Data Augmentation: ECE=$0.9\%$), and for CIFAR-10 when MLPs are ensembled (Pilot$_{\mathrm{MC}}$ \textit{a-aug}: ECE=$3.6\%$ vs Ensemble: ECE=$2.7\%$).
Nevertheless, in both cases these methods provide lower test set accuracy and NLL compared to Pilot \textit{a-aug}. 

Figure \ref{fig:calib_plot} shows reliability diagrams for our Pilot models and our various regularisation baselines.
Appendix \ref{app:mc_reliability} shows reliability diagrams for our Pilot models where we are running the classifier with samples $\tilde{a}$, as well as our baselines for model uncertainty estimation, MC dropout and ensembles.
Pilot activation augmentation, the top left of each sub figure, consistently gives well calibrated models with high reliability.

Appendix \ref{app:ent} contains histograms of the entropy of the predictive distributions over the test set for our Pilot models against baselines for CNNs and MLPs.
The Pilot models generally produce predictions with greater uncertainty.

\section{Discussion}
Overall Pilot \textit{a-aug} results in well-calibrated classifiers that exhibit superior performance to any of the baselines.
Calibration is important in many real work classification problems, where asymmetric loss shifts the prediction from $\argmax_y p(y|x)$.
Furthermore, it is compact: it does not require multiple forward pass samples, as in MC dropout, or training and storing a variety of models, as in ensembling, to produce accurate and calibrated predictions.

That generalising data augmentation to activations gives a modelling benefit is not unreasonable.
In a deep net, data and activations have similar interpretations.
For instance, the activations of the penultimate layer are features on which one trains logistic regression, so augmenting in this space has the same flavour as doing data augmentation for logistic regression.

Importantly, Pilot \textit{a-aug} outperforms our noise addition and substitution baselines meaning that our model performance cannot be solely attributed to noise injections in the fashion of \citet{Poole2014}. 
Note that we also include results in Appendix \ref{app:noisestop} for the additive noise baselines where we do not propagate gradients through inserted activation values, thus mimicking the exact conditions under which Pilot operates. These methods outperform their counterparts with propagated gradients, particularly for SVHN, which in itself is an interesting observation. 
Nevertheless, Pilot \textit{a-aug} also outperforms additive noise baselines with this design choice. 

One could view our model as performing a variety of \textit{transfer learning}: the samples from a larger generative model are used to train a smaller discriminative model (which is also the source of the training data for the larger model).
In addition, our approach constitutes a form of \textit{experience replay} \citep{Mnih2015nature}: the generative model learns a posterior over the discriminative model's activations by amortising inference over previous training steps, not solely relying on the $a=f_\Psi(x)$ from the current training iteration.

Pilot$_{\mathrm{MC}}$ models exhibit a small degradation in accuracy and NLL relative to their Pilot counterparts. 
Note that MC Dropout experiences a larger drop in accuracy relative to Dropout, and that Pilot$_{\mathrm{MC}}$ \textit{a-aug} still outperforms other uncertainty estimate models in terms of accuracy and NLL. 
Nevertheless, Pilot$_{\mathrm{MC}}$ models lead to a noticeable increase in ECE (save for CIFAR-10 MLPs), especially for CNNs. This could be due to the fixed variance of our decoder (set to 0.1) which may degrade model performance at test time.  
Our MC models, at the expense of calibration, can offer uncertainty estimates with little degradation to test-set accuracy and NLL.
Refining the calibration of our MC models is an area for further research. 

We are pleased to present here \textit{Pilot}, an effective new regularisation strategy for deep nets, and we hope this stimulates further research into regularisers that are trained alongside their discriminator.

\clearpage
\bibliographystyle{humannat}
\bibliography{references}
\newpage
\appendix
\onecolumn
\section{Normal-Gamma Hyperprior on z}
\label{app:hyperprior}

We place a Normal-Gamma hyperprior on z
and define $p_\theta(z|a_{1-b},b)$ as: 

\begin{align}
p_\theta(z, \mu_\psi, \sigma_\psi|a_{1-b},b) = \mathcal{N}(z|\mu_\psi,\sigma^2_\psi)\mathcal{N}(\mu_\psi|0,\sigma_\mu) \mathrm{Gamma}(\sigma_\psi|2,\sigma_\sigma)
\end{align}

As in \cite{Vetrov2019} this adds $-\frac{\mu_\psi}{2\sigma_\mu^2}$ and $\sigma_\sigma(log(\sigma_\psi) - \sigma_\psi)$ as penalties to the model log likelihood. 
For our MLP encoders and decoders we found that empirically $\sigma_\mu=10$ and $\sigma_\sigma=1$, combined with gradient clipping, promote stability and allow for the model to converge.

\section{Noisy Addition Results with No Gradient Propagation}
\label{app:noisestop}

\begin{table*}[h]
\vspace{1em}
  \caption{Test set accuracy and mean per-datapoint negative log-likelihood ($\mathrm{NLL}= -\ELBO_{\mathrm{xent}}$) for convolutional neural networks and 2-hidden-layer MLPs (with 1024 hidden units) trained on CIFAR-10 and SVHN with noisy addition (Add) baselines, where we do not propagate gradients through inserted activation values.}
  \label{table:noise_grad_results}
  \centering
    \setlength\tabcolsep{1.9pt} 
  \begin{tabular}{rcccc}
    \toprule
    \multicolumn{5}{c}{\textbf{CNN models}}\\
    \toprule    Model & Acc$(D^\mathrm{CIFAR10}_\mathrm{test})$ & Acc$(D^\mathrm{SVHN}_\mathrm{test})$& $\mathrm{NLL}(D^\mathrm{CIFAR10}_\mathrm{test})$ &  $\mathrm{NLL}(D^\mathrm{SVHN}_\mathrm{test})$\\
    \midrule
    \midrule
    Add  \textit{a-aug}     & $0.660$ & $0.863 $  & $1.51 $ & $ 0.56$ \\
    Add  \textit{a-drop}    & $0.632$ & $0.850$  & $1.42$ & $ 0.61$ \\
    \midrule
    \bottomrule
    \toprule
    \multicolumn{5}{c}{\textbf{MLP models}}\\
    \toprule
    Model & Acc$(D^\mathrm{CIFAR10}_\mathrm{test})$ & Acc$(D^\mathrm{SVHN}_\mathrm{test})$& $\mathrm{NLL}(D^\mathrm{CIFAR10}_\mathrm{test})$ &  $\mathrm{NLL}(D^\mathrm{SVHN}_\mathrm{test})$\\
    \midrule
    
    \midrule
    Add  \textit{a-aug}         & $0.578$ & $0.8436$  & $2.61$ & $ 0.74$\\
    Add  \textit{a-drop}     & $0.569$ & $0.8461$  & $2.23$ & $ 0.99 $\\

    \midrule
    \bottomrule
  \end{tabular}
\end{table*}

\newpage
\section{Reliability Diagrams for Sampling Methods}
\label{app:mc_reliability}

\begin{figure*}[h]
    \centering
    \subfloat[CIFAR10 - CNN]{{ \includegraphics[width=0.7\textwidth]{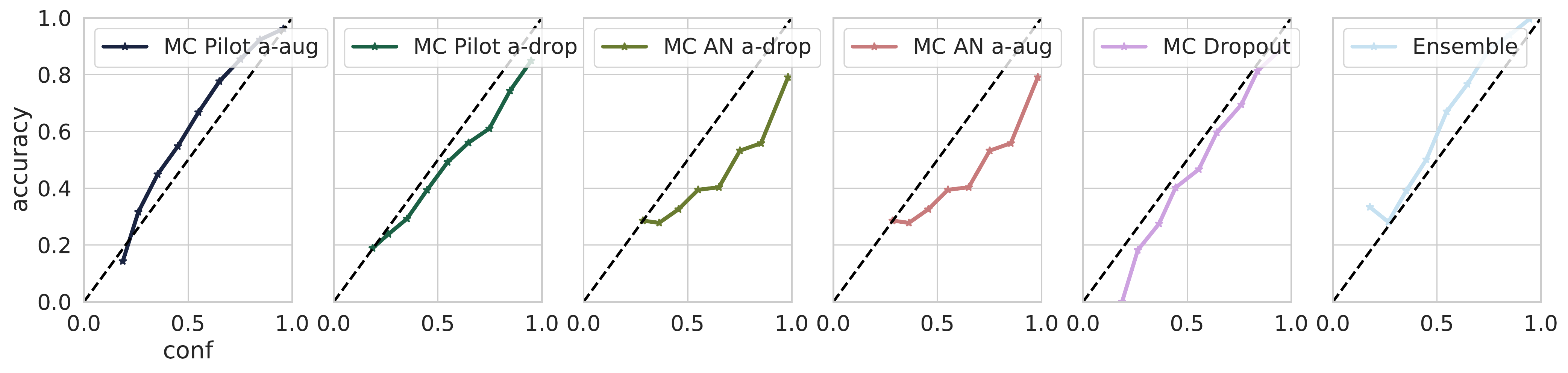} }}%
    \vspace{0mm}
    \subfloat[SVHN - CNN]{{ \includegraphics[width=0.7\textwidth]{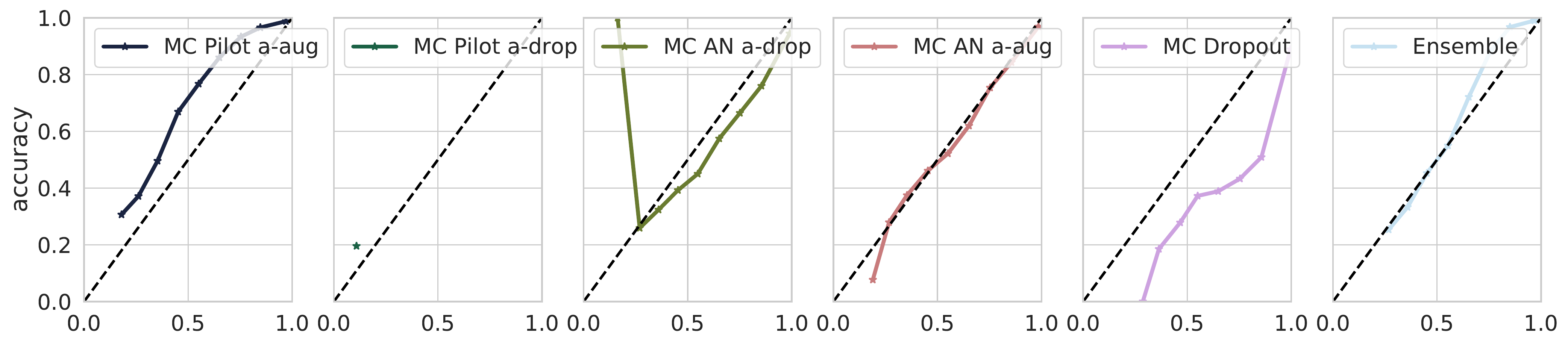} }}%
    \vspace{0mm}
    \subfloat[CIFAR10 - MLP]{{ \includegraphics[width=0.7\textwidth]{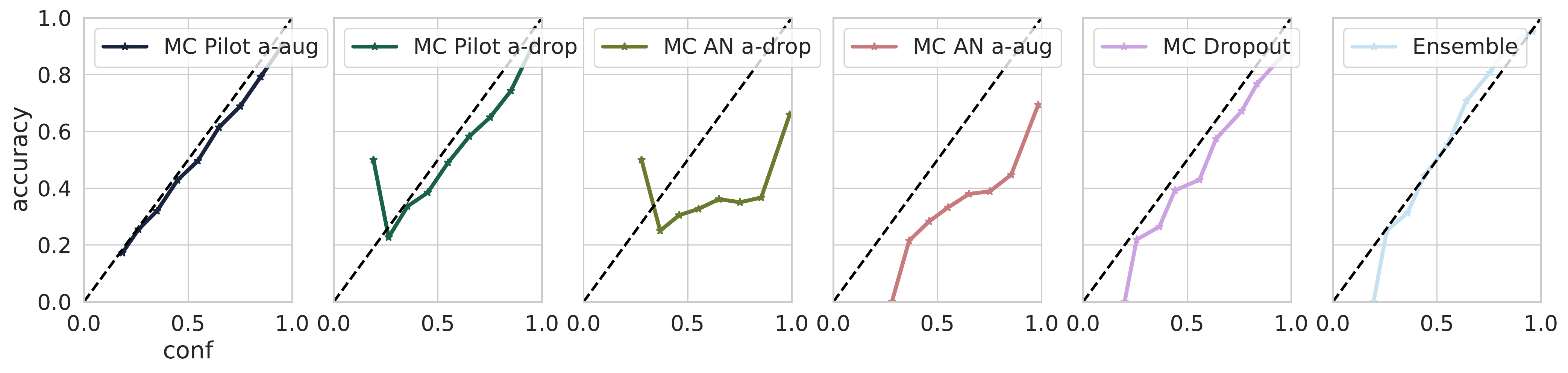} }}%
    \vspace{0mm}
    \subfloat[SVHN - MLP]{{ \includegraphics[width=0.7\textwidth]{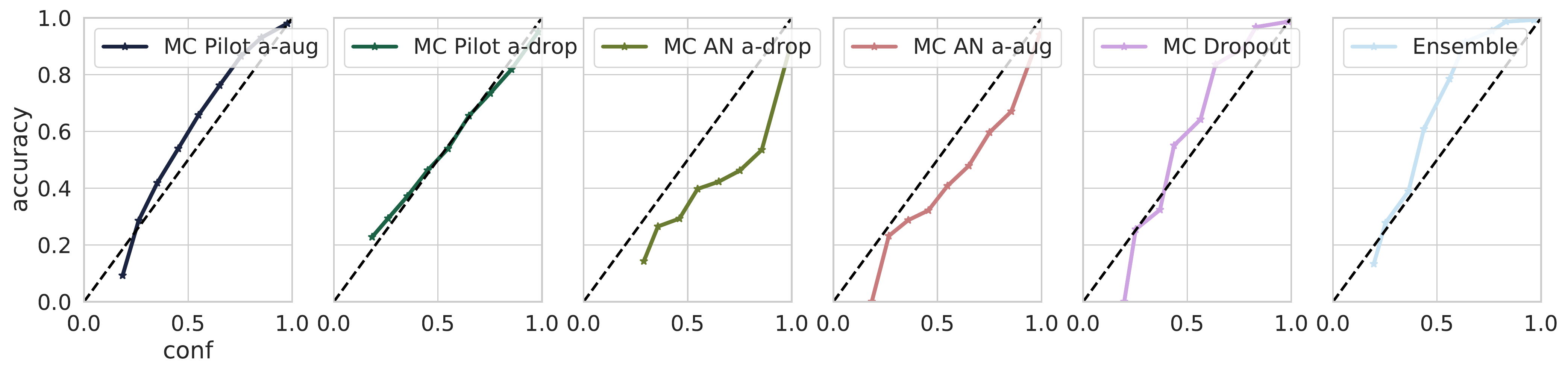} }}%
    \caption{Reliability diagrams for SVHN and CIFAR10 test sets for different for methods giving approximate model uncertainty.}
    \label{fig:calib_mc_plot}
\end{figure*}

\newpage
\twocolumn
\section{Entropy Histograms}
\label{app:ent}
\begin{figure}[h]
    \centering
    \subfloat[CIFAR10 - MLP]{{ \includegraphics[width=7cm]{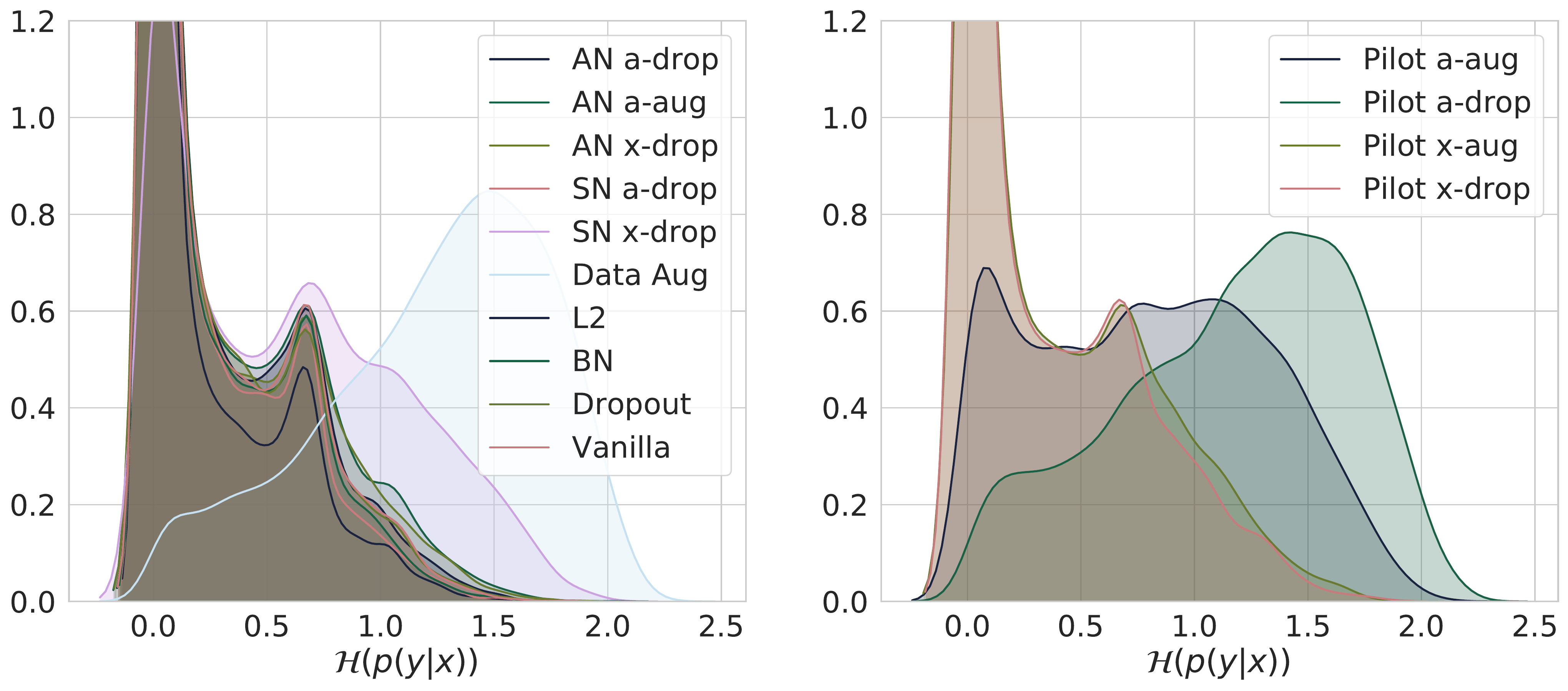} }}%
    \\
    \subfloat[SVHN - MLP]{{ \includegraphics[width=7cm]{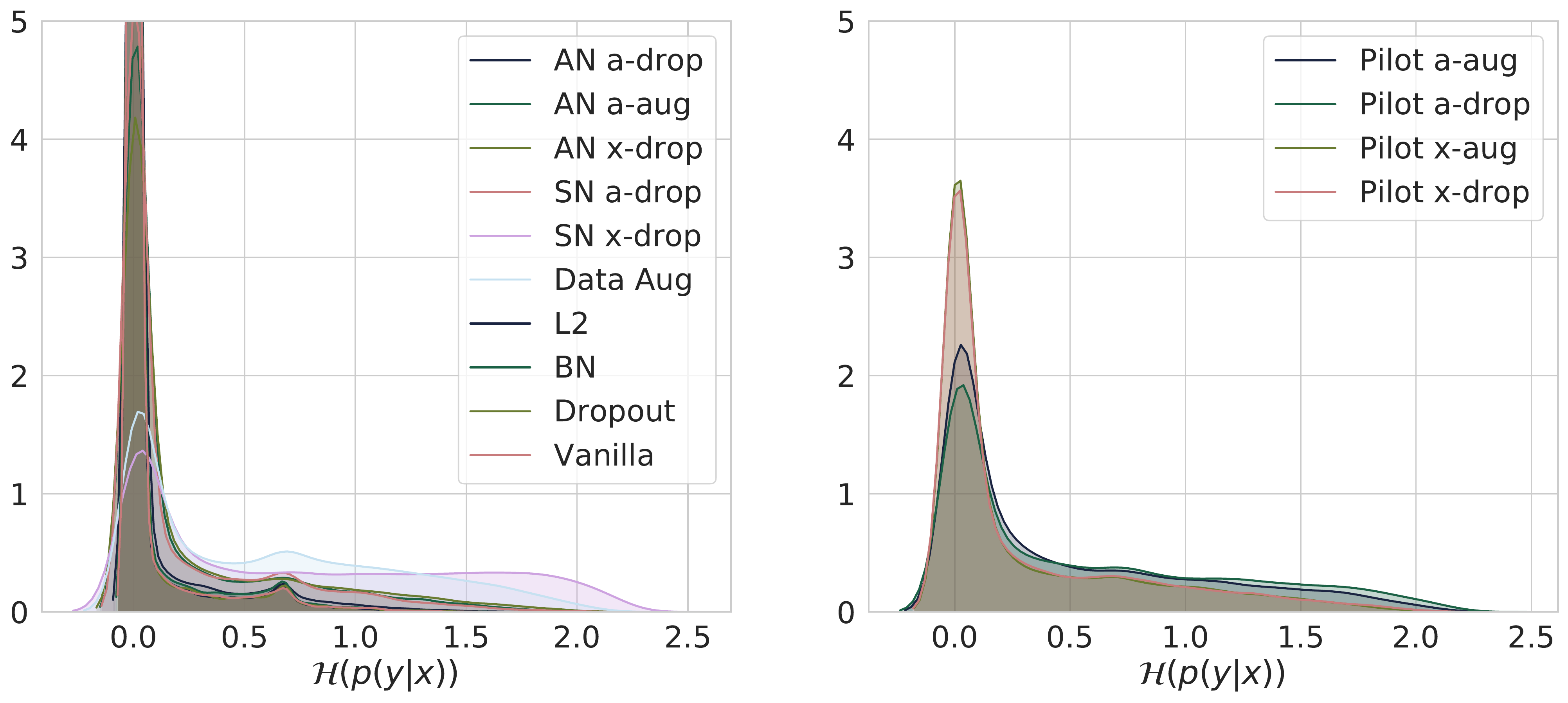} }}%
    \\
    \subfloat[CIFAR10 - CNN]{{ \includegraphics[width=7cm]{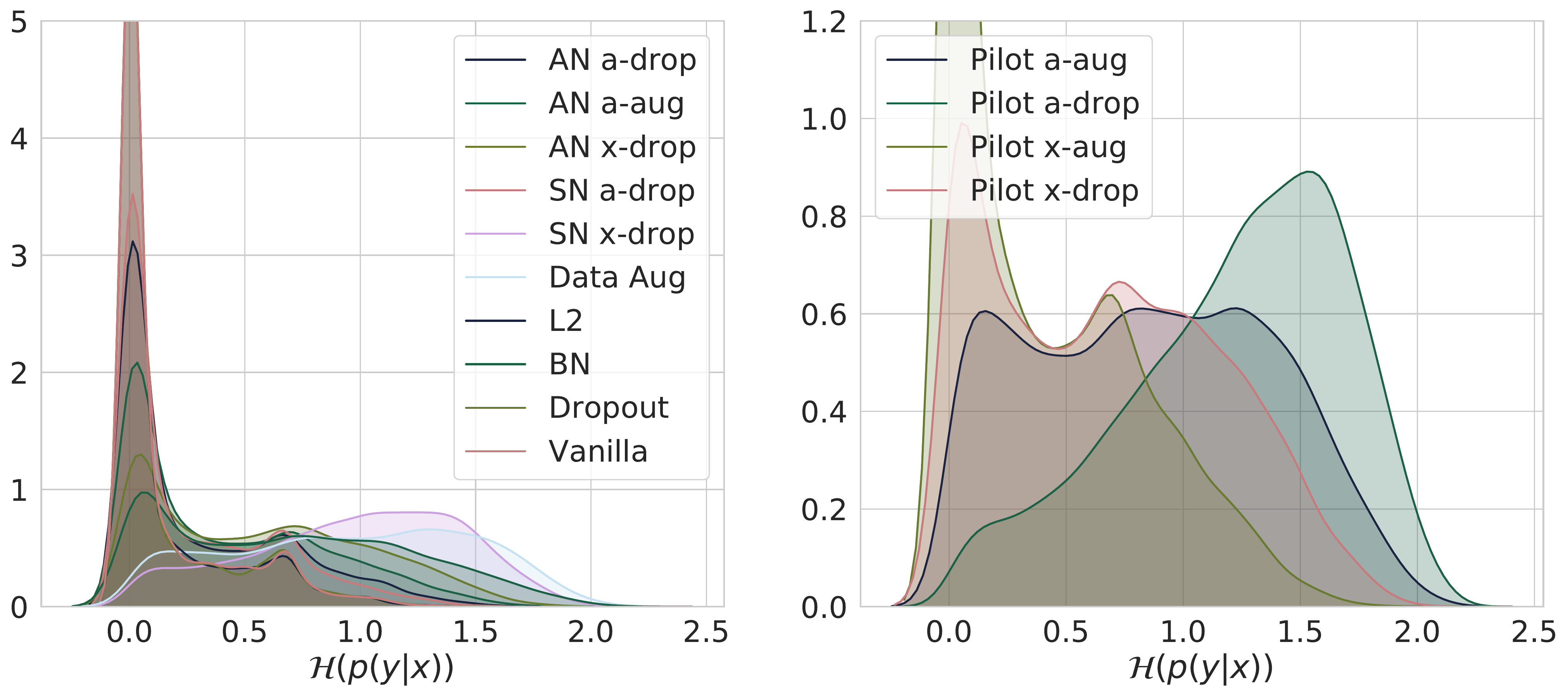} }}%
    \\
    \subfloat[SVHN - CNN]{{ \includegraphics[width=7cm]{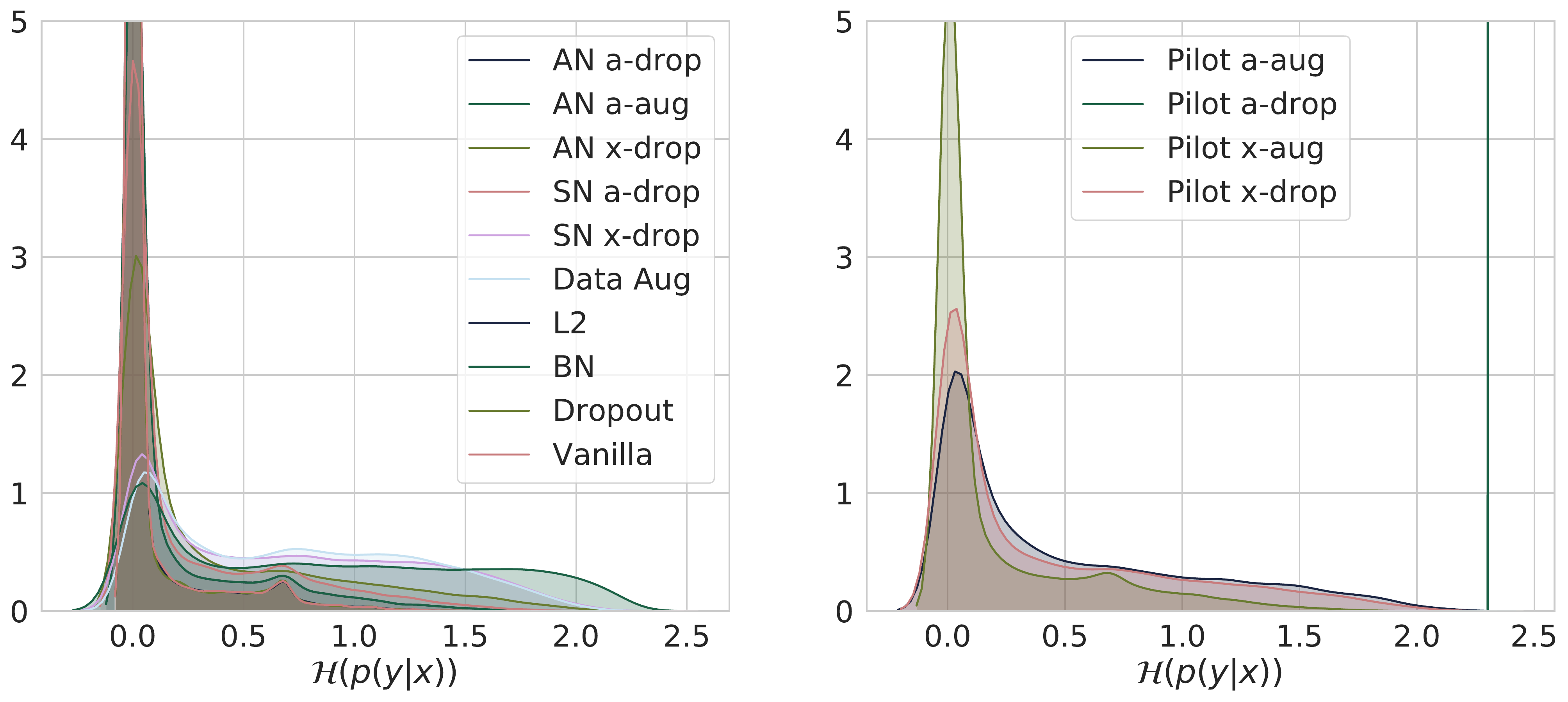} }}%
    \caption{Histogram of $\ent(p_\Psi(y|x))$ over the SVHN and CIFAR10 test sets for different regularisation methods on our MLPs and CNNs.}
    \label{fig:ent_hist_appendix}
\end{figure}
\newpage
\begin{figure}[h]
    \vspace{0.7cm}
    \centering
    \subfloat[CIFAR10 - MLP]{{ \includegraphics[width=7cm]{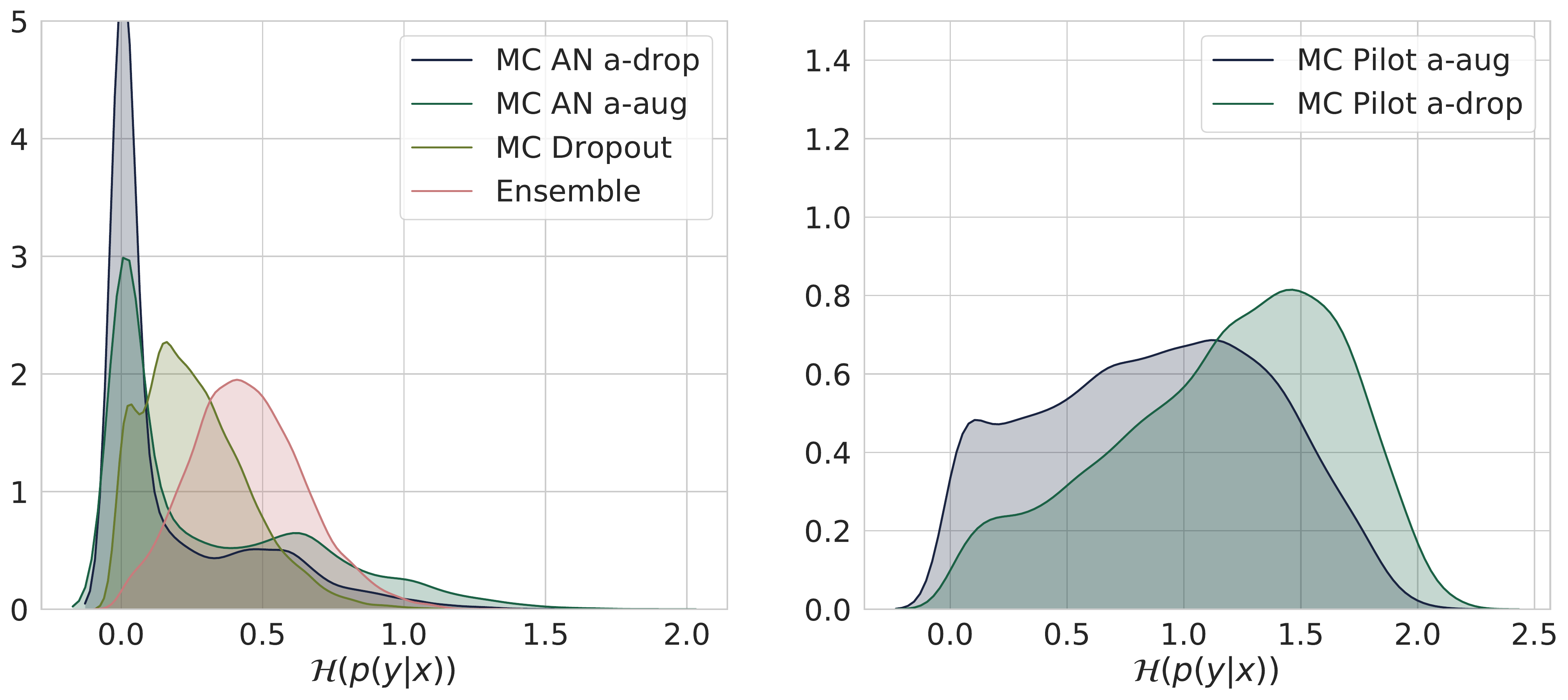} }}%
    \\
    \subfloat[SVHN - MLPN]{{ \includegraphics[width=7cm]{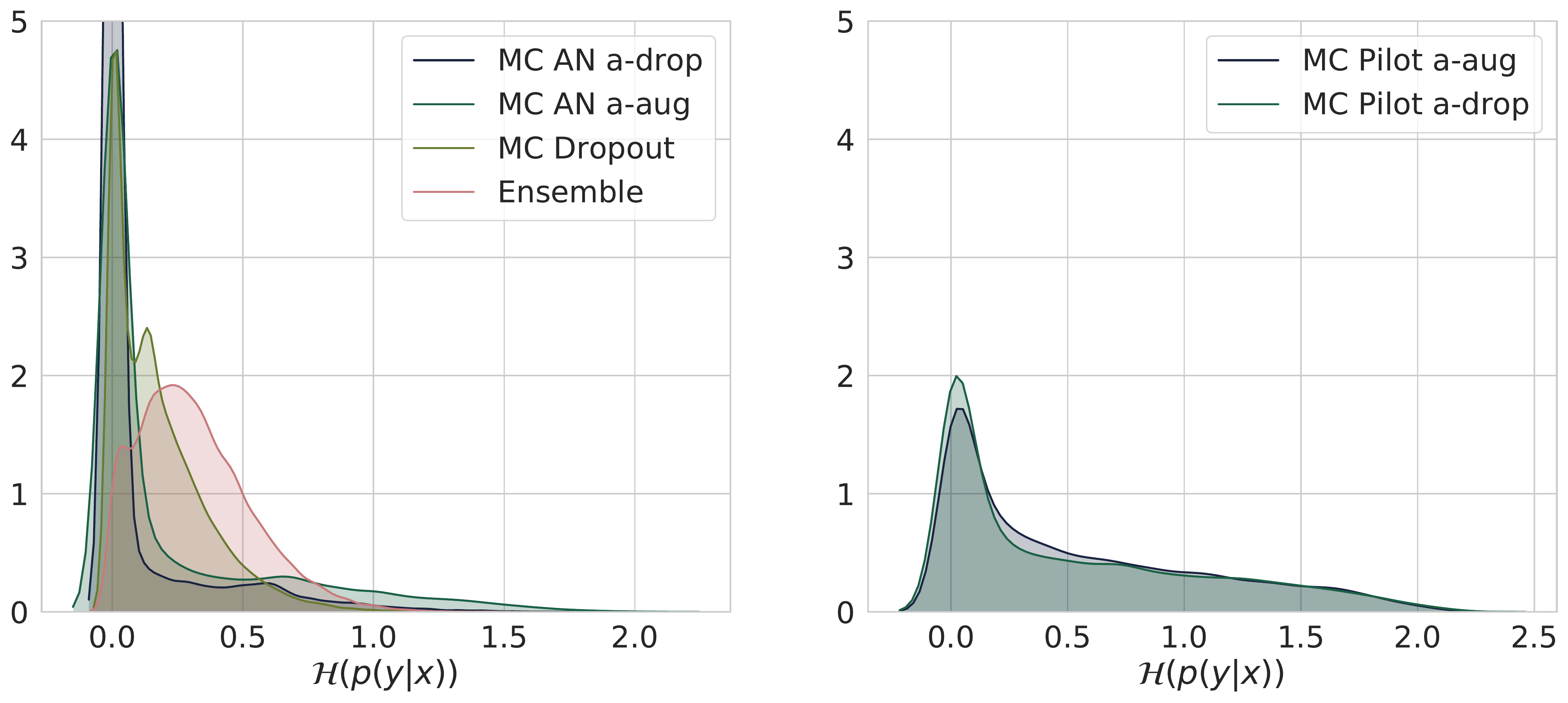} }}%
    \\
    \subfloat[CIFAR10 - CNN]{{ \includegraphics[width=7cm]{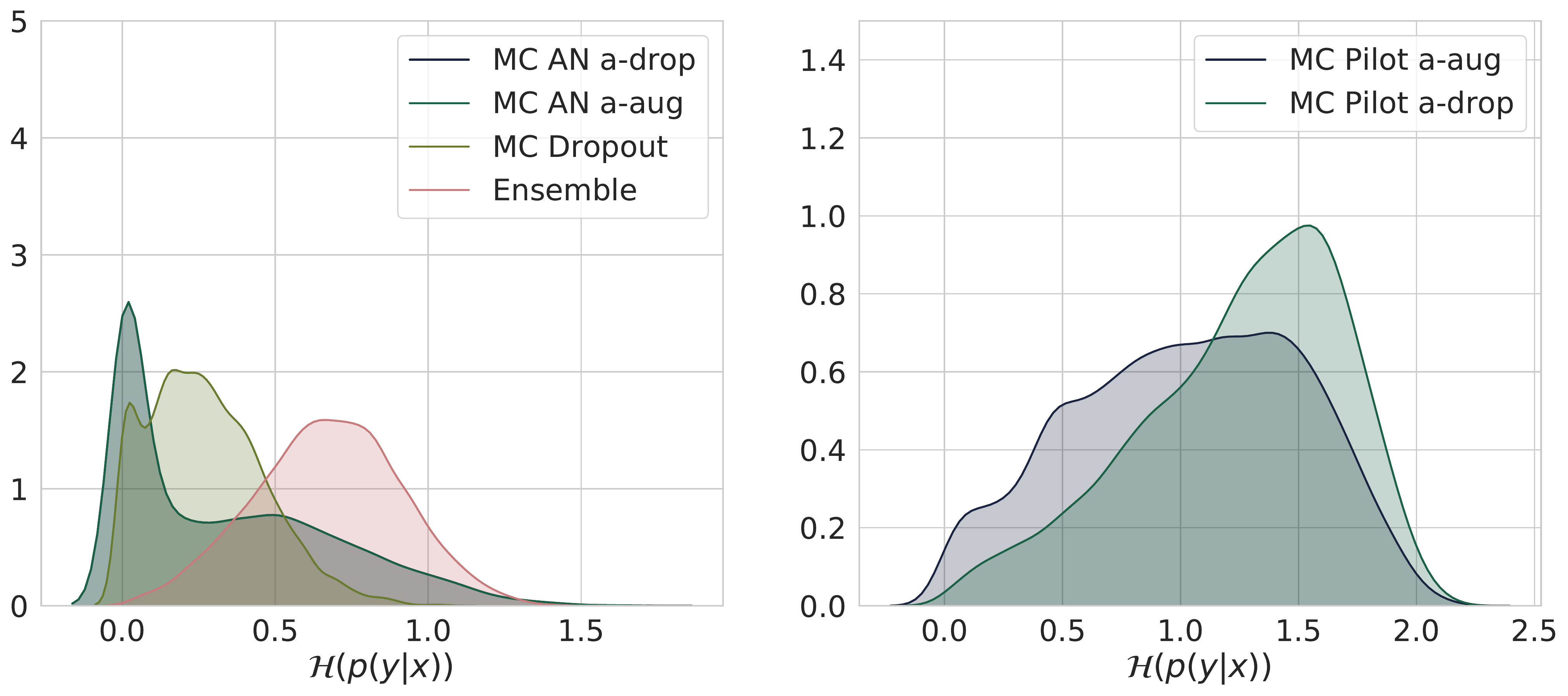} }}%
    \\
    \subfloat[SVHN - CNN]{{ \includegraphics[width=7cm]{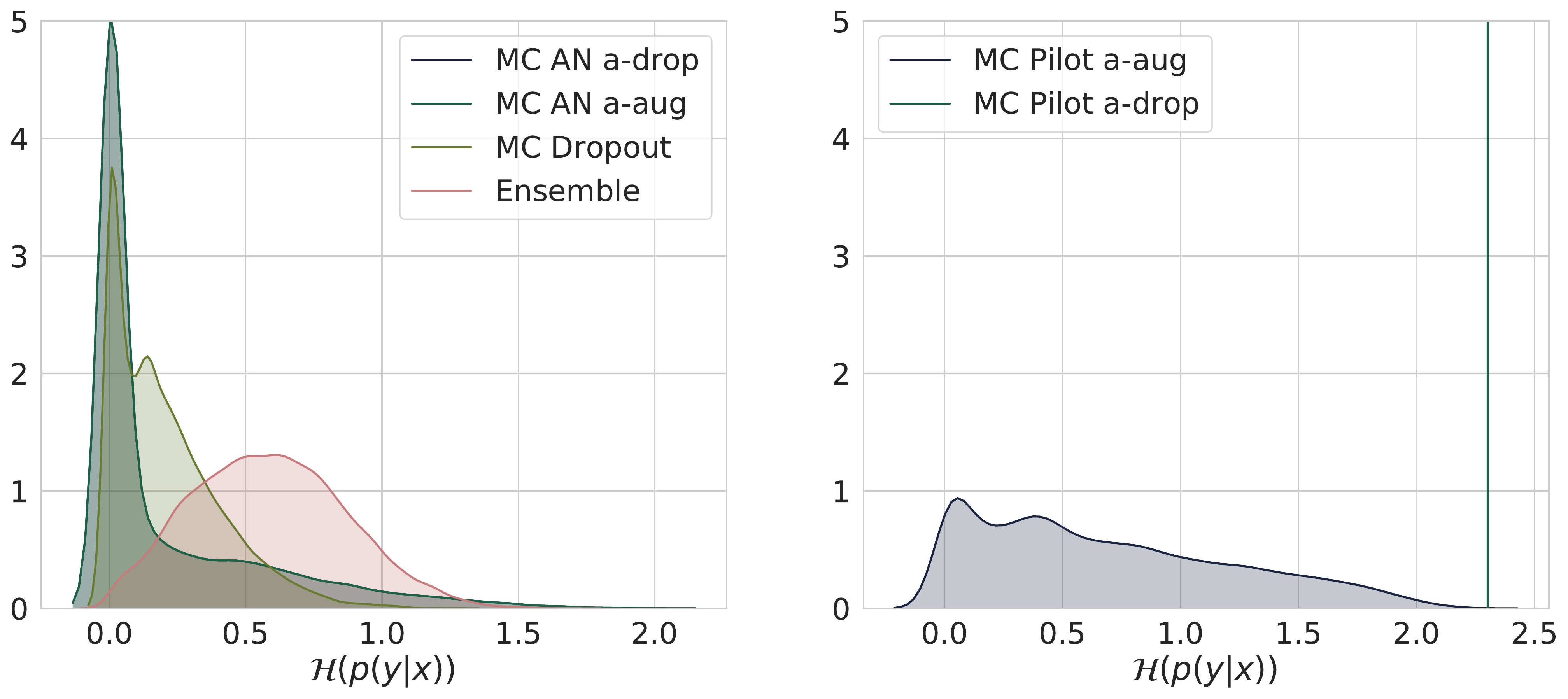} }}%
    \caption{Histogram of $\ent(p_\Psi(y|x))$ over the SVHN and CIFAR10 test sets for our MLPs and CNNs for different regularisation methods that give model uncertainty estimates}
    \label{ent_hist_appendix_mc}
\end{figure}

\newpage
\onecolumn
\section{Small MLP Results}
\begin{table}[h!]
\vspace{1em}
  \caption{Test set accuracy, mean per-datapoint negative log-likelihood ($\mathrm{NLL}= -\ELBO_{\mathrm{xent}}$) and ECE for our small MLP (2 hidden layers each with 512 neurons) trained on CIFAR-10 and SVHN with different regularisation methods}
  \label{table:small_mlp_results}
  \centering
\setlength\tabcolsep{1.9pt} 
  \begin{tabular}{lcccccc}
    \toprule
    Model & Acc$(D^\mathrm{CIFAR10}_\mathrm{test})$ & Acc$(D^\mathrm{SVHN}_\mathrm{test})$& $\mathrm{NLL}(D^\mathrm{CIFAR10}_\mathrm{test})$ &  $\mathrm{NLL}(D^\mathrm{SVHN}_\mathrm{test})$ & ECE$(D^\mathrm{CIFAR10}_\mathrm{test})$ & ECE$(D^\mathrm{SVHN}_\mathrm{test})$ \\
    \midrule
    \midrule
    Pilot \textit{aug}          & $\bm{0.562 \pm 0.012}$ & $0.830 \pm 0.014$ & $\bm{1.31 \pm 0.02}$ & $\bm{0.72 \pm 0.09}$ & 0.052 & 0.063 \\
    Pilot \textit{drop}         & $0.481 \pm 0.005$ & $0.66 \pm 0.055$ & $1.51 \pm 0.02$ & $ 0.99 \pm 0.01$ & $\bm{0.041}$ & $\bm{0.037}$\\
    \midrule
    Sub \textit{drop}           & $0.440 \pm 0.003$ & $ 0.747 \pm 0.022$  & $5.80\pm 0.15$ & $ 1.83 \pm 0.15$ & 0.451 & 0.184\\
    Add \textit{drop}           & $0.546 \pm 0.015$ & $0.810  \pm 0.004$  & $7.30\pm 0.03$ & $ 2.52 \pm 0.26$ & 0.399 & 0.144 \\
    Sub \textit{aug}            & $0.475 \pm 0.008$ & $0.7415 \pm 0.017$ & $2.80 \pm 0.01$ & $ 0.93 \pm 0.03$ & 0.333 & 0.044 \\
    Add \textit{aug}            & $0.542 \pm 0.013$ & $0.825  \pm 0.012$  & $4.80\pm 0.28$ & $ 1.05 \pm 0.06$ & 0.370 & 0.107 \\
    \midrule
    Dropout                     & $0.542 \pm 0.009$ & $ \bm{0.845 \pm 0.003}$  & $2.71  \pm 0.24$&$ 2.81 \pm 0.29$ & 0.310 & 0.136  \\
    $L_2$, $\lambda=0.1$        & $0.557 \pm 0.007$ & $ 0.832 \pm 0.001$  & $3.00 \pm 0.06$ & $ 2.63 \pm 0.17$ & 0.334 & 0.143  \\
    Batch norm                  & $0.560 \pm 0.016$ & $ 0.832 \pm 0.006$  & $2.90 \pm 0.55$ & $ 1.93 \pm 0.10$ & 0.475 & 0.553  \\
    Data Aug                    & $0.493 \pm 0.001$ & $ 0.678 \pm 0.008$  & $2.12 \pm 0.01$ & $ 1.26 \pm 0.02$ & 0.281 & 0.150  \\
    \midrule
    \bottomrule
  \end{tabular}
\end{table}

\end{document}